\documentclass[10pt,twocolumn,letterpaper]{article}

\usepackage{cvpr}              % To produce the CAMERA-READY version
\usepackage[dvipsnames]{xcolor}

\definecolor{cvprblue}{rgb}{0.21,0.49,0.74}
\usepackage[pagebackref,breaklinks,colorlinks,citecolor=cvprblue]{hyperref}

\usepackage{float}
\usepackage{stfloats}
\usepackage{enumerate}

\usepackage{svg}
\usepackage{amsmath}

\graphicspath{{Images/}}

\def\paperID{27} % *** Enter the Paper ID here
\def\confName{CVPR}
\def\confYear{2024}

\title{EucliDreamer: Fast and High-Quality Texturing for 3D Models with Stable Diffusion Depth}

\author{Cindy Le\\
Columbia University\\
{\tt\small xl2738@columbia.edu}
\and
Conrui Hetang\\
Carnegie Mellon University\\
{\tt\small congruihetang@gmail.com}
\and
Chendi Lin\\
Carnegie Mellon University\\
{\tt\small chendil@alumni.cmu.edu}
\and
Ang Cao\\
University of Michigan\\
{\tt\small ancao@umich.edu}
\and
Yihui He\\
Carnegie Mellon University\\
{\tt\small he2@alumni.cmu.edu}
}

\usepackage{listings}
\usepackage{xcolor}

\begin{document}
\twocolumn[{%
\maketitle
\begin{center}
    \centering
    \includegraphics[width=0.99\textwidth]{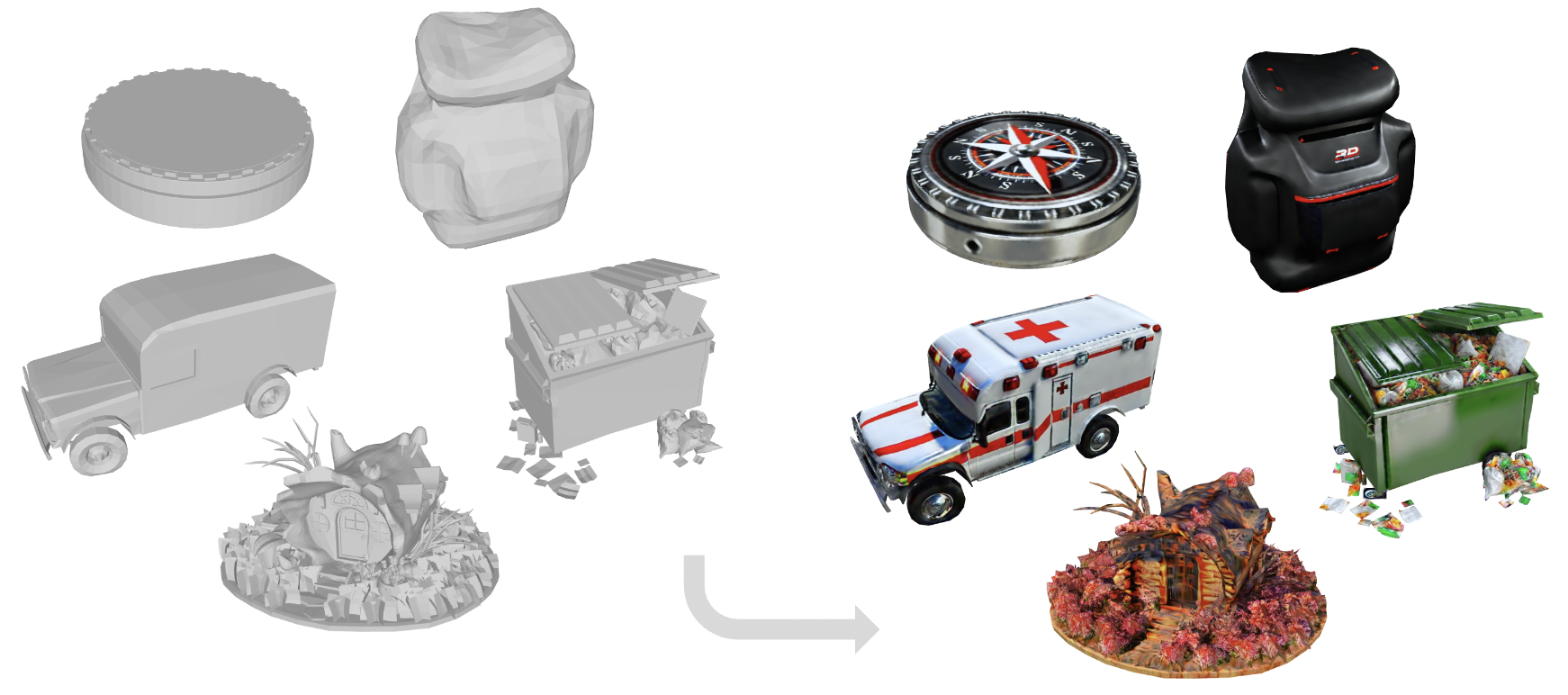}
    \captionof{figure}{A showcase of objects textured by EucliDreamer.}
    \label{fig:cover}
\end{center}
}]

\begin{abstract}

This paper presents a novel method to generate textures for 3D models given text prompts and 3D meshes. Additional depth information is taken into account to perform the Score Distillation Sampling (SDS) process~\cite{poole2022dreamfusion} with depth conditional Stable Diffusion~\cite{Rombach_2022_CVPR}.  We ran our model over the open-source dataset Objaverse~\cite{deitke2022objaverse} and conducted a user study to compare the results with those of various 3D texturing methods. We have shown that our model can generate more satisfactory results and produce various art styles for the same object. In addition, we achieved faster time when generating textures of comparable quality. We also conduct thorough ablation studies of how different factors affect generation quality, including sampling steps, guidance scale, negative prompts, data augmentation, elevation range, and alternatives to SDS.

\end{abstract}  

\section{Introduction}
\label{sec:intro}

3D modeling is widely used in the media, gaming, and education industries to create characters and environments, realistic product visualizations and animations, and interactive simulations and learning materials. According to Mordor Intelligence, the global market size of 3D modeling and animation has reached 6 billion dollars~\cite{marketsize}. Meanwhile, manual creation of high-quality 3D models is still challenging. The lengthy and labor-intensive process involves creating the 3D mesh, 2D textures, and the correct mapping between them. Any common object (e.g. a house, a tree, etc.) can take an experienced artist several days to make. 

We aim to automate the texturing process, which takes a considerable amount of time. The problem is formulated as: given a 3D mesh representing an object, generating a reasonable coloring for all the points on its surface, defined by a 2D image and a 2D-to-3D mapping.

Ever since the last decade, researchers have been looking for methods to assist 3D creation. Earlier methods applied Generative Adversarial Networks (GAN)~\cite{wu2017learning, chen2018text2shape} and Contrastive Language-Image Pre-Training (CLIP)~\cite{jain2022zeroshot,Mohammad_Khalid_2022, michel2021text2mesh}. Recent years have witnessed the wide adoption of diffusion-based methods for high-quality 3D content generation~\cite{poole2022dreamfusion}.

\begin{figure}
    \centering
    \includegraphics[width=0.4\textwidth]{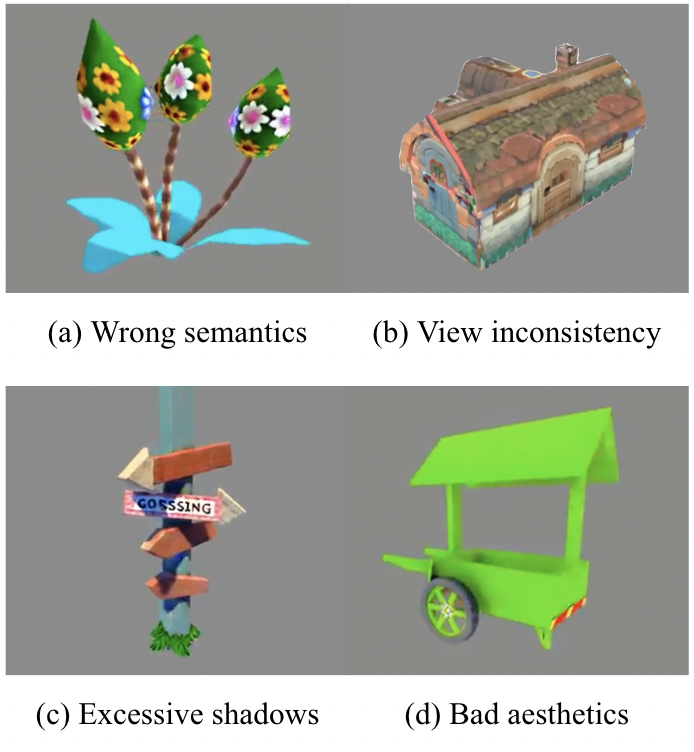}
    \caption{\textbf{Problematic textures} that we generated using existing Diffusion-based methods. (a) The flower model has blue leaves and flower patterns over the flower buds due to an incorrect understanding of the model. (b) The house has different colors and styles for each wall. (c) The sign has shadows on the texture, which should only result from rendering. (d) The cart has an oversaturated color of bright green and does not look pleasant. }
    \label{fig:problem}
\end{figure}

Existing works for 3D texturing still face many problems including poor quality, low diversity, and bad view consistency. Figure~\ref{fig:problem} showcases four typical issues encountered in generated textures, underscoring the existing shortcomings in this area.

\begin{itemize}

\item \textbf{Wrong semantics.} If the 3D mesh represents a car, the corresponding texture should give the tires a dark rubber color and not cover the window with colorful patterns.

\item \textbf{View inconsistency.} Because popular texturing methods derive multi-views from a single-view image, the resulting 3D object may have different content or color themes. This happens often when texturing buildings, cars, and other objects that have symmetric structures.

\item \textbf{Excessive light reflections and shadows.} Despite the industry having various standards and practices for 3D texturing, it is generally preferable that they do not contain shadows before rendering~\cite{Disney}. Thus, a texture with heavy shadows becomes unusable.

\item \textbf{Bad color tones.} Sometimes a 3D texture does not fall into any of the prior categories but the appearance is bad in a subjective way. This usually happens when the texture violates color principles. They may combine two colors that do not match, or have a color with too high saturation.

\end{itemize}

3D modeling data is scarce by its nature and many are owned by private entities, which puts pressure on algorithms to use existing datasets efficiently.

This paper presents a novel method that adopts Stable Diffusion depth~\cite{rombach2021highresolution} to generate 3D texturing that yields better quality. Our work presents the following contributions: (1) a pioneering approach that integrates Stable Diffusion depth within the Score Distillation Sampling (SDS) process for enhancing 3D texturing tasks; (2) comprehensive experimental analyses and a user study that collectively demonstrates significant improvements in inference speed.

\section{Related Work}
\label{sec:related}

\subsection{Early attempts of 3D generation}

Inspired by the works on text-guided generation in the 2D space~\cite{yu2022vectorquantized, karras2019stylebased}, many early attempts for 3D generation adopted GAN~\cite{goodfellow2014generative}. Jiajun Wu et al.~\cite{wu2017learning} proposed a 3D-GAN to generate 3D objects from a probabilistic space. Similarly, when Contrastive Language-Image Pre-Training (CLIP)~\cite{radford2021learning} was proposed for joint embedding in 2021, the CLIP-based works on 2D~\cite{patashnik2021styleclip, crowson2022vqganclip} generation naturally inspired explorations in 3D~\cite{jain2022zeroshot, Mohammad_Khalid_2022, michel2021text2mesh}.

Dreamfields~\cite{jain2022zeroshot} proposed a text-guided 3D
generation using Neural Radiance Field (NeRF)~\cite{mildenhall2020nerf}. It is an inverse rendering approach to create 3D models from a set of 2D images. 

\subsection{3D modeling with SDS and diffusion models}

In 2022, DreamFusion~\cite{poole2022dreamfusion} first proposed using a pre-trained 2D text-to-image model for text-guided 3D generation through differentiable rendering. Their core method, Score Distillation Sampling (SDS), samples uniformly in the parameter space from pre-trained diffusion models to obtain gradients that match given text prompts.

DreamFusion~\cite{poole2022dreamfusion} inspired many SDS-based improvements for 3D generation. Magic3D~\cite{lin2023magic3d} further improved the quality and the speed of 3D modeling by a two-step approach, which obtains a coarse model using a low-resolution diffusion, accelerating with a sparse 3D hash grid structure, and then interacting with a high-resolution latent diffusion model. Fantasia3d~\cite{chen2023fantasia3d} further disentangles geometry and appearance generation, supervised by the SDS loss.

Our method adopts SDS as well. We will also explore alternatives to SDS in the Ablation Studies ~\ref{sec:ablation}.

\subsection{3D texturing and depth information}

While much research focuses on generating a whole 3D model, such generated output can hardly be used in practice due to complicated industry-specific requirements. For example, 3D models in gaming are rendered in real-time, which limits the number of polygons and requires clear topological structures. 

In contrast, AI-powered 3D texturing is a more realistic goal. One notable research on texturing is TEXTure~\cite{richardson2023texture} which can generate a new texture and its surface-to-surface mapping given a new 3D mesh. It applies Stable Diffusion depth~\cite{rombach2021highresolution} but only as a simple mesh projection. Text2Tex~\cite{chen2023text2tex} proposes a powerful texuring method involving two stages, generation and refinement. It also incorporates depth information but does so through ControlNet~\cite{zhang2023adding}.

As a 3D texturing method, our EucliDreamer will directly use Stable Diffusion depth. We will compare it with the ControlNet depth in the Ablation Studies  ~\ref{sec:ablation}.

\subsection{View consistency for 3D modeling}

The NeRF~\cite{mildenhall2020nerf} and SDS process require multi-view images for a single object, which may not be accessible due to a lack of data. As a result, generating multi-views from a single image becomes a crucial step in high-quality texture generation. Such generations sometimes suffer poor consistency across different views, for example, a house with four walls all in different color patterns, or a car whose color is asymmetric.

Zero-1-to-3~\cite{liu2023zero1to3} proposes a method to synthesize images of an object from new angles through diffusion models. Other research works~\cite{shi2023mvdream, tang2023mvdiffusion, liu2023syncdreamer} also attempted to address the view consistency problem.

Different from Text2Tex~\cite{chen2023text2tex}, SDS enforces 3D consistency by interaction. we further improve consistency using a large batch size.

\section{Method}
\label{sec:method}

We present our novel method that, for the training loop with SDS loss, depth information is taken into account in addition to the color information.

\subsection{Depth conditioning}

\begin{figure*} [h!]
    \centering
    \includegraphics[width=0.8\textwidth]{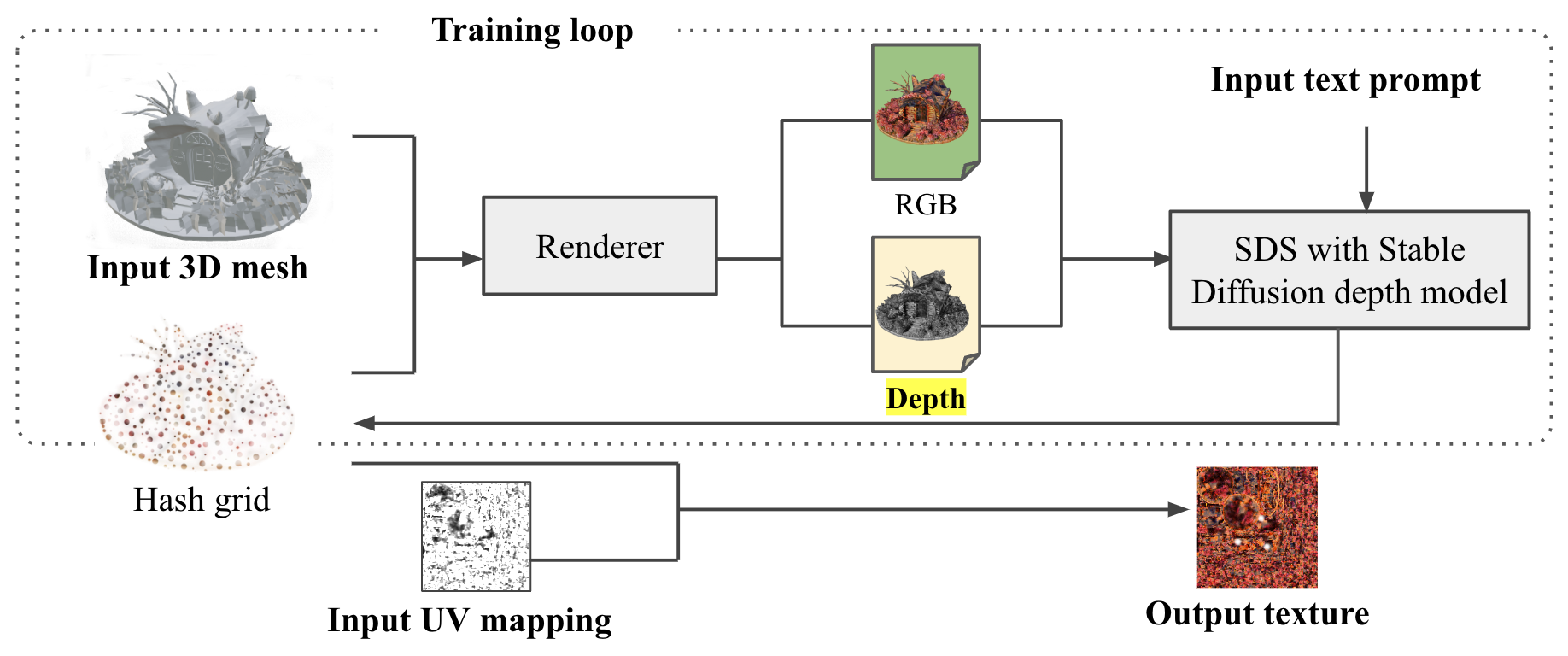}
    \caption{\textbf{Our method.} A render will transform the 3D model into a 2D view and extract layers of information. Then the SDS loss will loop back to update a hash grid. An additional layer of Stable Diffusion depth is added along with the RGB color layer.}
    \label{fig:method}
\end{figure*}

Our method, inspired by DreamFusion~\cite{poole2022dreamfusion}, is composed of the rendering stage and the SDS training stage. Different than previous methods, we add a depth layer of Stable Diffusion along with the RGB color layer.

Our model takes a 3D mesh that represents the model's shape as input and generates texture for it iteratively. The texture is represented as a hash grid~\cite{mueller2022instant}. Initially, the mesh is textured with random texture. On each iteration, the textured mesh is fed into a differentiable renderer~\cite{Laine2020diffrast} that renders RGB image and depth from multiple different angles. Subsequently, we use score distillation sampling (SDS) to guide the iteration of the texture. In SDS, a depth conditional diffusion model~\footnote{https://huggingface.co/stabilityai/stable-diffusion-2-depth} takes the RGB image as input, conditioned on text prompt and depth. The model keeps iterating over the hash grid and updating it through the same rendering and computation process guided by SDS loss. The hash grid is converted at the end of all iterations for popular 3D formats.

Input UV maps are optional. They are surface-to-surface mappings to guide the coloring of the meshed 3D model. In practice, it is usually in the interest of users to input the UV to convert the NeRF~\cite{mildenhall2020nerf} view so the resulting texture output is easy to edit further.

\subsection{Datasets}

Our rendering step is based on the open-sourced Stable Diffusion 2~\cite{rombach2021highresolution, Rombach_2022_CVPR} that can generate and modify images based on text prompts. We run our model over objects from Objaverse~\cite{deitke2022objaverse}, a large open-sourced dataset of objects with 800K+ textured models with descriptive captions and tags. 

\subsection{Experiments}

We conducted two sets of experiments to show the power of our EucliDreamer.

For the first set of experiments, we run our model over several 3D meshes from the open-sourced Objeverse to observe the generation quality. We use the best parameters from our Ablation Studies~\ref{sec:ablation}. In addition to the text input to other models, we add a group of the same keywords to all the prompts in our model, which can be viewed as part of our algorithm. 

For the second set of experiments, we run our model over fixed 3D meshes with custom text prompts and observe the style variations. We select some 3D models from Objaverse~\cite{deitke2022objaverse} that are typical to represent a category, such as cars, furniture, buildings, etc. For each model, we try different combinations of prompts to generate textures in various art styles.

\subsection{Evaluation}

We compare our results with those of Text2Tex~\cite{chen2023text2tex}, CLIPMesh~\cite{Mohammad_Khalid_2022}, and Latent-Paint~\cite{metzer2022latentnerf}. We conducted a user study of 28 participants including CS researchers, artists, and practitioners from the gaming industry. They are asked to rank the results in terms of generation quality.

\begin{figure*} [h!]
    \centering
    \includegraphics[width=0.8\textwidth]{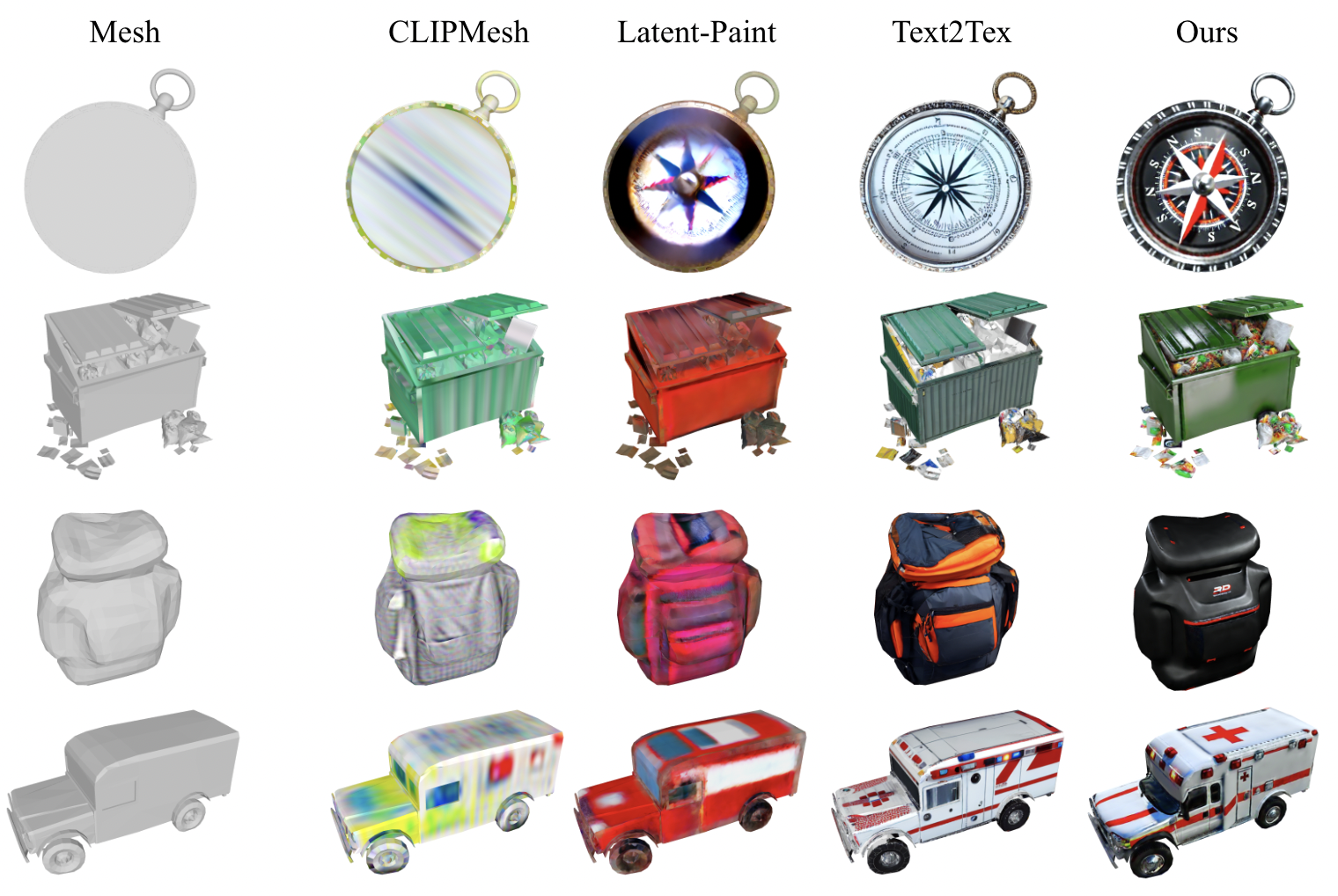}
    \caption{\textbf{Comparisons with previous texturing methods.} Four objects are used for illustration from Objaverse~\cite{deitke2022objaverse}. The rendering performance of the first three methods, CLIPMesh, Latent-Paint, and Text2Tex are discussed in \cite{chen2023text2tex}. Overall, the examples demonstrate a clear win of Text2Tex~\cite{chen2023text2tex} and our method against the baselines methods~\cite{Mohammad_Khalid_2022, metzer2022latentnerf} in terms of clarity and level of detail.}
    \label{fig:result}
\end{figure*}

\begin{figure*}
    \centering
    \includegraphics[width=0.8\textwidth]{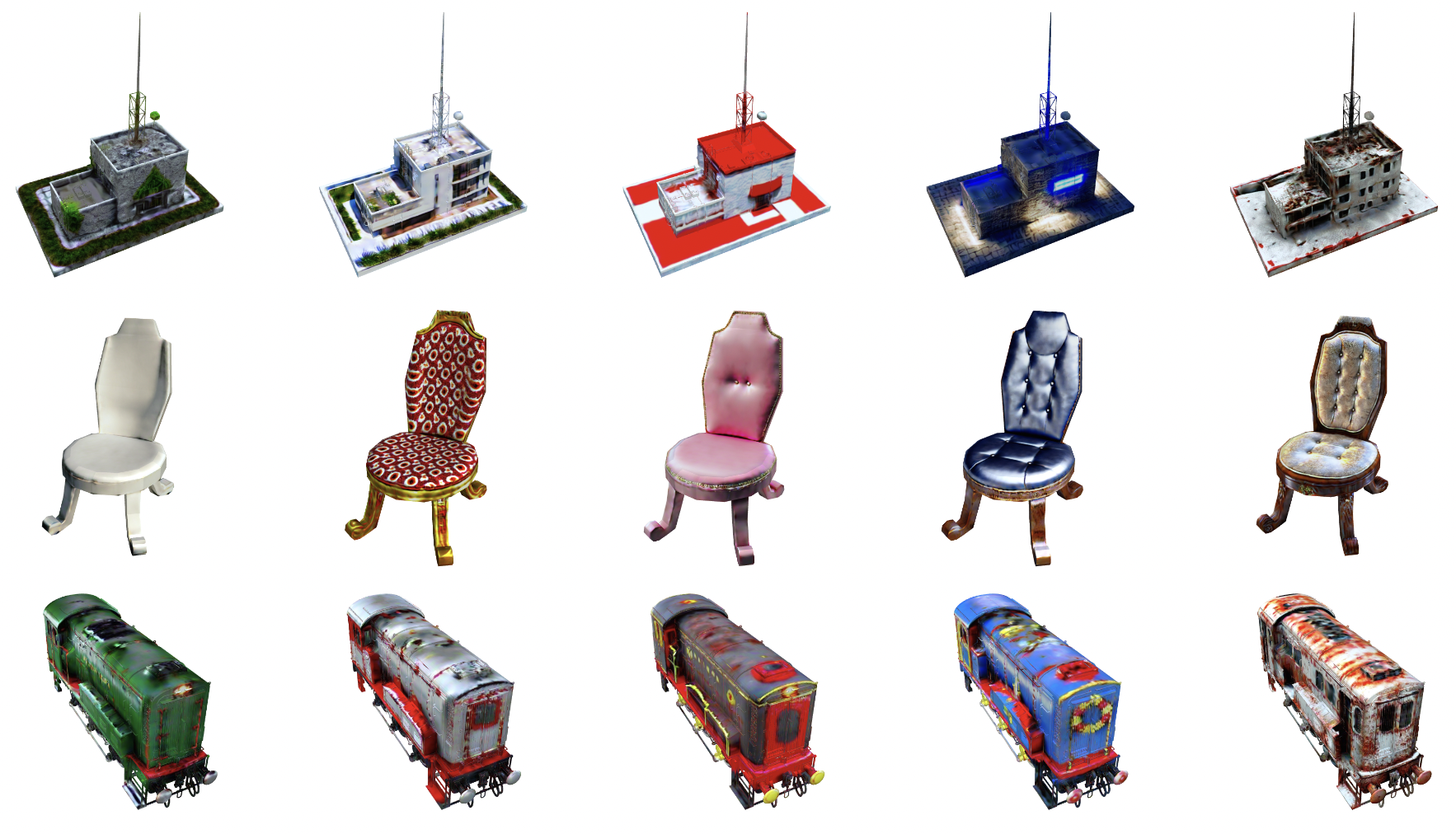}
    \caption{Objects textured by EucliDreamer in various styles, by different input text prompts.}
    \label{fig:styles}
\end{figure*}

\section{Ablation Studies}
\label{sec:ablation}

For model understanding and parameter selection, we conduct a series of experiments for EucliDreamer with different settings.

\subsection{SDS vs. VSD}

\begin{figure*}
    \centering
    \includegraphics[width=0.7\textwidth]{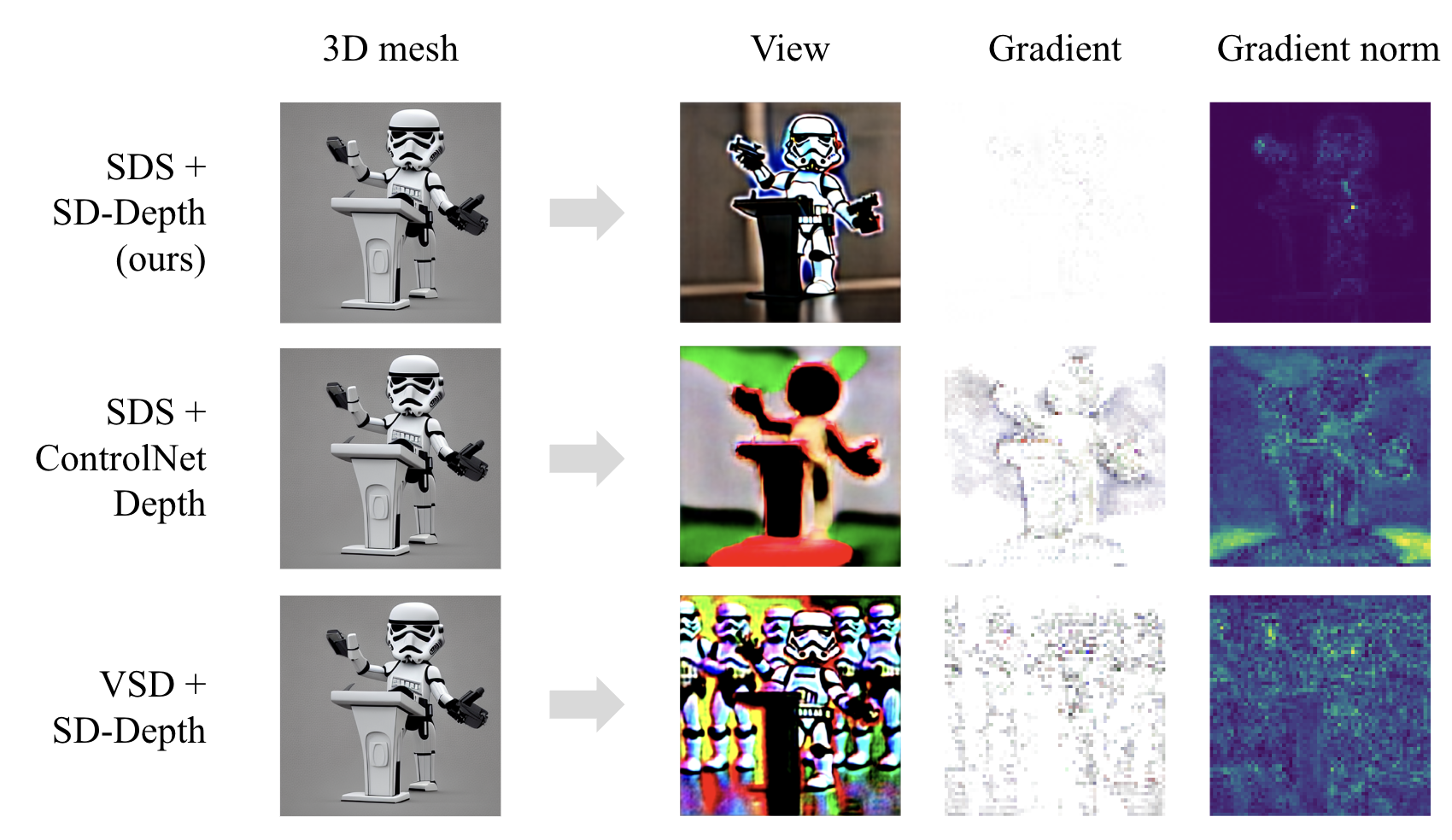}
    \caption{\textbf{Comparisons among multiple depth techniques.} The SDS and ControlNet depth combination generates a view with missing details of the character. The VSD~\cite{wang2023prolificdreamer} and Stable Diffusion depth combination leads to multiple characters and noisy gradient results. Overall, Our method with SDS and Stable Diffusion depth generates the best gradient for the given mesh.}
    \label{fig:controlnet}
\end{figure*}

SDS~\cite{poole2022dreamfusion} is notoriously known for its lack of diversity and overly smoothed texture. This is because SDS is mean-seeking. Intuitively, a single prompt has multiple plausible generation results. SDS tends to create an average of them (mean seeking). Variational Score Distillation (VSD) is proposed in ProlificDreamer~\cite{wang2023prolificdreamer} to mitigate this problem by introducing a variational score that encourages diversity. However, in depth-conditioned texturing, there are fewer plausible generation results given a single prompt and depth. This leads to the diversity and details in our approach (mode seeking), even though VSD is not used. Through experiments shown in Figure~\ref{fig:controlnet}, our approach produces better results without VSD. Thus VSD in ProlificDreamer~\cite{wang2023prolificdreamer} is not required when Stable Diffusion depth is used.

\subsection{StableDiffusion depth vs. ControlNet depth}

The most straightforward way of depth-conditioned SDS is to use ControlNet~\cite{zhang2023adding}. However, empirically we found that the SDS gradient from ControlNet~\cite{zhang2023adding} is much noisier than SD-depth. SD-depth concatenates depth with latents. The SDS gradients flow to depth and latents separately, which is less noisy. As shown in Figure~\ref{fig:controlnet}, we demonstrate better generation results with Stable Diffusion depth than ControlNet depth. 

\subsection{Elevation range}
\label{sec:elevation}
Elevation range defines camera positions and angles at the view generation step. When it is set between 0-90 degrees, cameras face down and have an overhead view of the object. We find that fixed-angle cameras may have limitations and may miss certain angles, resulting in blurry color chunks at the surface of the object. One way to leverage the parameter is to randomize camera positions so all angles are covered to a fair level.

\subsection{Sampling min and max timesteps}

The sampling min and max timesteps refer to the percentage range (min and max value) of the random timesteps to add noise and denoise during SDS process. The minimum timestep sets the minimum noise scale. It affects the level of detail of the generated texture. The maximum timestep affects how drastic the texture change on each iteration or how fast it converges. While their values may affect quality and convergence time, no apparent differences were observed between different value pairs of sampling steps. In general, 0 and 1 should be avoided for the parameters.

\subsection{Hyperparameters}
\label{sec:hyperparameters}
Based on multiple sets of experiments, the hyperparameters are decided and listed in Supplementary Materials Section \ref{sec:suppl1}.

\subsection{With vs. without gradient clipping}

Intuitively gradient clipping can avoid some abnormal updates and mitigate the Janus problem, shown in Debiased Score Distillation Sampling (D-SDS)~\cite{hong2023debiasing}. In the context of texture generation, we do not observe Janus problem due to the fact that the mesh is fixed. In our experiments, there's no noticeable benefit of gradient clipping. 

\subsection{Guidance scale}
\label{sec:guidance}
The guidance scale specifies how close the texture should be generated based on input text prompts. We find that the guidance scale affects diversity and saturation. Even with the depth conditional diffusion model, a high guidance scale like 100 is still required. 

\subsection{Negative prompts}
\label{sec:negative}
Negative prompts in Stable Diffusion can fix some artifacts caused by SDS. Adding negative prompts like "shadow, green shadow, blue shadow, purple shadow, yellow shadow" will help with excessive shadows.

\subsection{Data augmentation}
\label{sec:augmentation}

\begin{figure}
    \centering
    \includegraphics[width=0.45\textwidth]{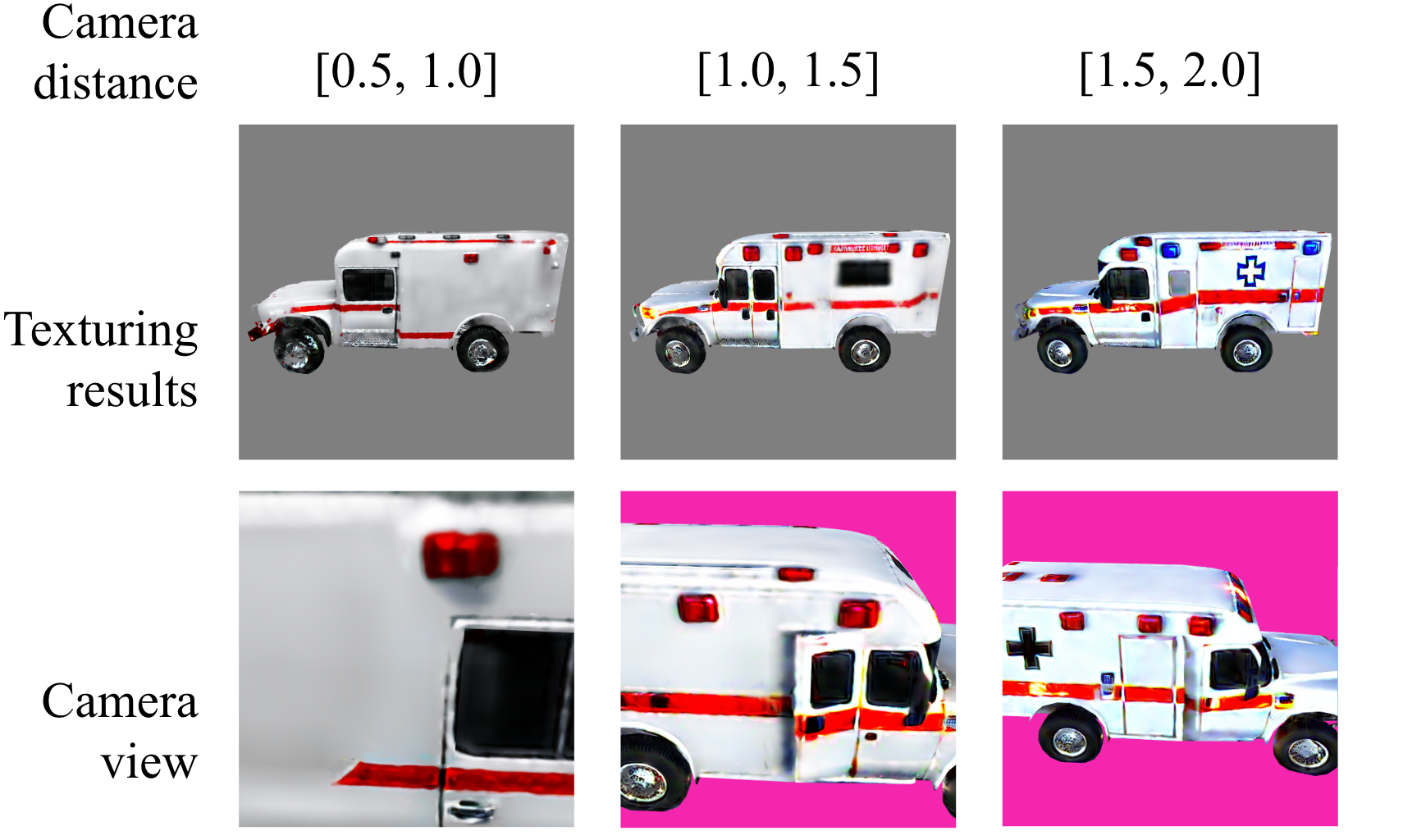}
    \caption{\textbf{Comparisons among camera distance values.} The one with [1.0, 1.5] has the highest quality.}
    \label{fig:camera}
\end{figure}

We found that adjusting the camera distance can improve both resolution and generation quality. We experimented with three different ranges of camera distance, shown in Figure~\ref{fig:camera}. The distance range of [1.5, 2.0] produces the optimal outcomes regarding the quality of textures and colors. This can be attributed to its ability to simulate viewpoints most representative of conventional photography, similar to the usual perspectives from which an ambulance is observed.

\subsection{Image-to-texture using Dreambooth3D}
\begin{figure}
    \centering
    \includegraphics[width=0.45\textwidth]{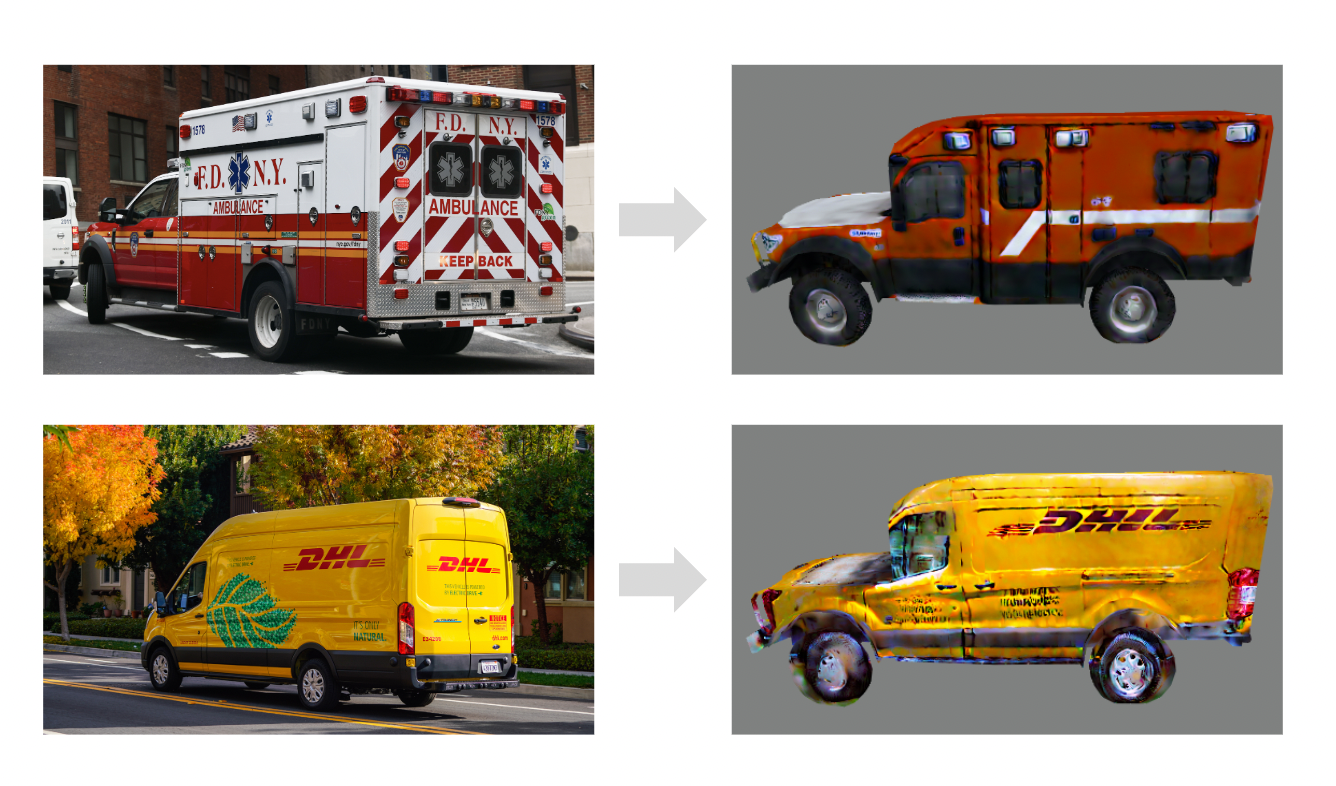}
    \caption{\textbf{Image-to-texture generation using Dreambooth3D~\cite{wang2023prolificdreamer}.} The generated textures have the same color tones as the input images.}
    \label{fig:dreambooth}
\end{figure}

Dreambooth3D~\cite{raj2023dreambooth3d} was initially proposed as an extension of DreamFusion~\cite{poole2022dreamfusion} to enable image-to-3D shape generation. We partially finetuned Stable Diffusion depth with the user-provided image(s), generated textures with finetuned stable-diffusion-depth from step 1, fine-tuned Stable Diffusion depth again with outputs from step 2, and then generated final results with finetuned stable-diffusion-depth from step 3. Shown in Figure~\ref{fig:dreambooth}, combined with Dreambooth3D~\cite{raj2023dreambooth3d}, our approach generates satisfactory textures given a 3D mesh and a single image as inputs. 

\section{Results}
\label{sec:results}

We adopt the best setup from Ablation Studies~\ref{sec:ablation} and conduct the main experiments. The results are exciting, as the generated textures are of good quality, achieve faster inference time with better aesthetics, and have various colors and styles through different text prompts.

\subsection{Benchmark objects from Objaverse}

The benchmark objects generated from our first set of experiments look realistic, detailed, and of high quality. The subtle bright areas of the front surface of the dumpster and the compass increase the level of detail, a trait seen in objects that apply a special technique named texture baking. The colored patterns of the compass are symmetric with straight outlines, which indicates a high generation quality.

\subsection{User study on generation quality} We selected 28 participants for the user study who are engineers, AI researchers, artists, and other game practitioners, mostly based in the U.S. About half of them have prior experience using a 3D modeling tool like Unity or Blender.

In the study, we present four sets of objects: an ambulance, a dumpster, a compass, and a backpack, via online questionnaires. Each set contains four textures generated by CLIPMesh, Latent-Paint, and Text2Tex, as reported by Text2Tex~\cite{chen2023text2tex}, and by our method. The participants are asked to rank the results in terms of generation quality of their definition, but not their preferences of colors. 

\begin{figure}[htbp]
    \centering
    \includegraphics[width=0.5\textwidth]{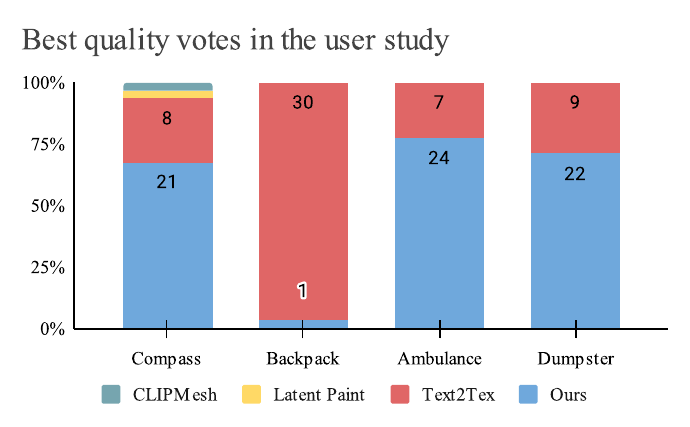}
    \caption{\textbf{Vote distribution for best quality.} Ours was selected as the best for most of the objects.}
    \label{fig:user1}

    \centering
    \includegraphics[width=0.5\textwidth]{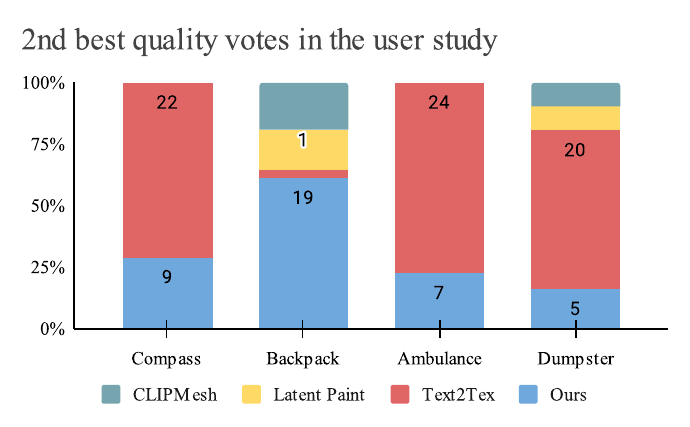}
    \caption{\textbf{Vote distribution for second best quality.} SDS-based methods overall received more votes than the CLIP-based ones.}
    \label{fig:user2}
\end{figure}

The results are shown in Figure~\ref{fig:result}. Most participants selected their top 2 choices from either SOTA Text2Tex~\cite{chen2023text2tex} or ours. For the ambulance and dumpster objects, the votes are unanimously toward our generated textures. For the compass, about two-thirds voted for ours. The votes for the backpack are diverse for the second-best quality object. 

Overall, the study indicates a high acceptance of the textures generated by our model.

\subsection{Faster inference time}

\begin{figure}
    \centering
    \includegraphics[width=0.5\textwidth]{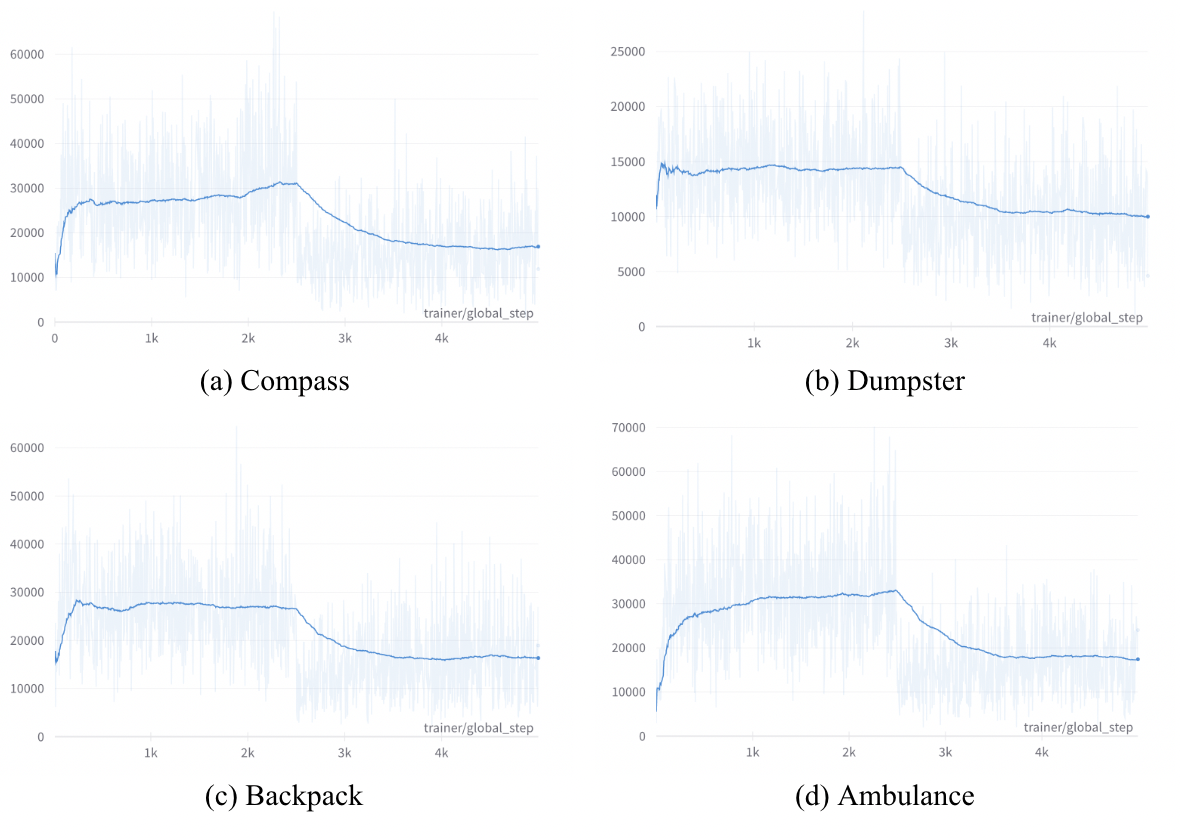}
    \caption{\textbf{The SDS loss throughout the iterations.} It takes around 4300 steps to converge and always has a sudden drop after 2500 steps.}
    \label{fig:loss}
\end{figure}

Figure~\ref{fig:loss} shows the SDS loss throughout the training steps. In general, it takes EucliDreamer around 4300 steps to converge the loss. Note that for every object, the loss drops drastically after 2500 steps, and the quality improvement is visually confirmable. In contrast, under identical experimental conditions and using the same set of hyperparameters, DreamFusion~\cite{poole2022dreamfusion} requires over 10,000 steps to converge. For both EucliDreamer and DreamFusion, with batch size of 8, our throughput is 2 iterations per GPU per second. Thus, the runtime of EucliDreamer is about one half of that of DreamFusion. This comparison highlights the superior convergence speed and computational efficiency of our method.

\subsection{Better aesthetics}

\begin{figure*}
    \centering
    \includegraphics[width=0.8\textwidth]{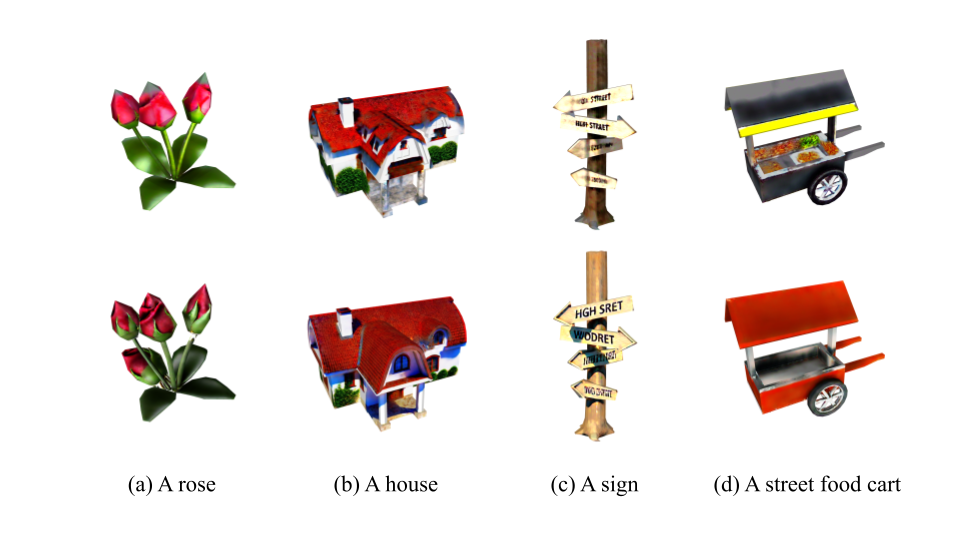}
    \caption{\textbf{Fixed problematic textures} using EucliDreamer. The top row does not include depth conditioning, and the bottom row does. (a) The flower model with correct semantics. (b) The house with consistent styles and themes. (c) The sign with correct shadows. (d) The cart with vivid colors. }
    \label{fig:problem_fix}
\end{figure*}

In Section~\ref{sec:intro}, we identified prevalent challenges in texturing quality, including semantics, view inconsistency, the handling of shadows, and maintaining accurate color tones. Our methodology addresses these concerns by generating textures that are significantly more realistic. Figure \ref{fig:problem_fix} demonstrates the substantial enhancements made to the four scenarios initially depicted in Figure~\ref{fig:problem}, utilizing identical prompts and 3D models using our method, with or without Stable Diffusion depth. The improvements are evident in the accurate and detailed semantics, refined textures, and well-rendered shadows, all while achieving a high degree of consistency across different viewpoints, especially with depth conditioning. The integration of Stable Diffusion depth into our model significantly enriches the generated textures with high-quality details, underscoring the considerable progress our method represents over prior works.

\subsection{Various styles}

Our second set of experiments focuses on the diversity of the generated textures. By providing input text prompts of the model content and styles, one can use EucliDreamer to generate 3D textures for a given 3D mesh in certain styles. For example, a prompt of "a building, damaged, dirty, 3D rendering, high-quality, realistic" will give the modeled building a rusty feel and walls in pale white, as shown in Figure~\ref{fig:styles}.

From an artistic perspective, art styles are subjective feelings resulting from combinations of color tones, brush strokes, lighting and shadows, and more. We will leave readers to judge the quality and creativity of the AI-generated textures.

\section{Discussions}
\label{sec:dicussions}

Through the experiments, we have proved that our method with Stable Diffusion conditioning can generate higher-quality textures in a shorter time. This is as expected, for the following reasons.

In terms of quality, depth information further restricts the probability density, eliminating unnatural or impossible cases. In some cases, without the depth layer, the model may generate a texture for a truck that has human characters on the surface, or a tree pattern on the leaves for a tree model.

In terms of generation time, the time of the loss conversion is much shorter for two reasons. First, depth conditioning serves as a constraint that eliminates many possibilities. Second, only the surface of the object is computed. Both led to a smaller probability space that takes less time to search.

\subsection{Significance}

Through the experiments and the user study, we have proved that adding Stable Diffusion depth to the SDS training can greatly improve the quality and speed of 3D texturing. Also, VSD~\cite{wang2023prolificdreamer} as an alternative to SDS is not a must when Stable Diffusion depth is enforced.

\subsection{Limitations}

Our method, like many other SDS-based ones, requires that all surfaces of an object be visible for effective coloring.

\subsection{Future directions}

We as a team still have much to do around the current 3D generation pipeline. Due to the 2-dimensional nature of Stable Diffusion, light reflections and shadows are not well handled during the SDS step. We will address this problem from both data and algorithm perspectives.

In 3D modeling, there are still many problems yet to be solved. Besides AI-powered texture generation, we also hope to research more on generating 3D meshes, scenes, animation, etc. We hope our work can inspire more research in 3D texturing and eventually solve the hassle in the real world.

\nocite{KaolinLibrary,ravi2020accelerating,hertz2023delta,sella2023voxe,armandpour2023reimagine,an2023panohead,wu2023omniobject3d,zhang2023avatarverse}
{
    \small
    \bibliographystyle{ieeenat_fullname}
    \bibliography{main}
}

% WARNING: do not forget to delete the supplementary pages from your submission 
\clearpage
\maketitlesupplementary

\setcounter{page}{1}

% \onecolumn
\section{Implementation details}
\label{sec:suppl1}

To better demonstrate our approach and reproduce our results, we will publish our code on GitHub upon acceptance.

% https://github.com/TCXX/Eucli-SdDepth

\subsection{Parameter selection}
\label{sec:parameters}

For the main experiment, we selected the following parameters that performed the best in ~\ref{sec:ablation}. The Adam optimizer has a learning rate of 0.01.

\begin{table}[!ht]
    \centering
    \begin{tabular}{|l|l|}
    \hline
        \textbf{Parameter} & \textbf{Value} \\ \hline
        Batch size & 8 \\ \hline
        Camera distance & [1.0, 1.5] \\ \hline
        Elevation range & [10, 80] \\ \hline
        Sampling step & [0.02, 0.98] \\ \hline
        Optimizer & Adam \\ \hline
        Training steps & 5000 \\ \hline
        Guidance scale & 100 \\ \hline
    \end{tabular}
\end{table}

The prompt keywords below are added in addition to the original phrase of the object (e.g.  “a compass”).

\begin{verbatim}
prompt:
  "animal crossing style, a house,
  cute, Cartoon, 3D rendering, red
  tile roof, cobblestone exterior"
negative_prompt:
  "shadow, greenshadow, blue shadow,
  purple shadow, yellow shadow"
\end{verbatim}

The texture resolution is set to 512*512.

\subsection{Infrastructure}

ThreeStudio

(https://github.com/threestudio-project/threestudio)

is a unified framework for 3D modeling and texturing from various input formats including text prompts, images, and 3D meshes. It provides a Gradio backend with easy-to-use frontend with input configurations and output demonstration.

We forked the framework and added our approach with depth conditioning. A depth mask is added to encode the depth information. 

\begin{verbatim}

with torch.no_grad():
    depth_mask =
      self.pipe.prepare_depth_map(
        torch.zeros((batch_size, 3,
          self.cfg.resolution,
          self.cfg.resolution)),
        depth_map,
        batch_size=batch_size,
        do_classifier_free_guidance=True,
        dtype=rgb_BCHW.dtype,
        device=rgb_BCHW.device,
    )

\end{verbatim}

\subsection{Hardware}

We used a Nvidia RTX 4090 GPU with 24GB memory. This allows us to run experiments with a batch size up to 8 and dimension 512*512. In theory, a GPU with more memory will allow us to generate textures with higher definition and in better quality.

\section{More results from Ablation Studies}
\label{sec:suppl}

Due to the page limit, we show experimental results of elevation range (section~\ref{sec:elevation}), batch size (section~\ref{sec:parameters}), guidance scale (section~\ref{sec:guidance}), negative prompts (section~\ref{sec:negative}), and data augmentation (section~\ref{sec:augmentation}) here. The results support our conclusions in the Ablation Studies section~\ref{sec:ablation}.

\subsection{Learning Rate}
In the context of 3D texture generation, a large learning rate usually leads to faster convergence. A small learning rate creates fine-grained details. We found learning rate of 0.01 with the Adam optimizer is good for 3D texture generation. 

\subsection{Batch Size}
We set batch sizes to 1, 2, 4, and 8 respectively. Results show that a larger batch size leads to more visual details and reduces excessive light reflections and shadows. More importantly, large batch size increases view consistency. This is because gradients from multiple views are averaged together and the texture is updated once. A batch size of 8 is enough from our observation. A batch size of more than 8 does not lead to a significant reduction in the number of iteration steps. 

\begin{figure}[h!]
    \centering
    \includegraphics[width=0.45\textwidth]{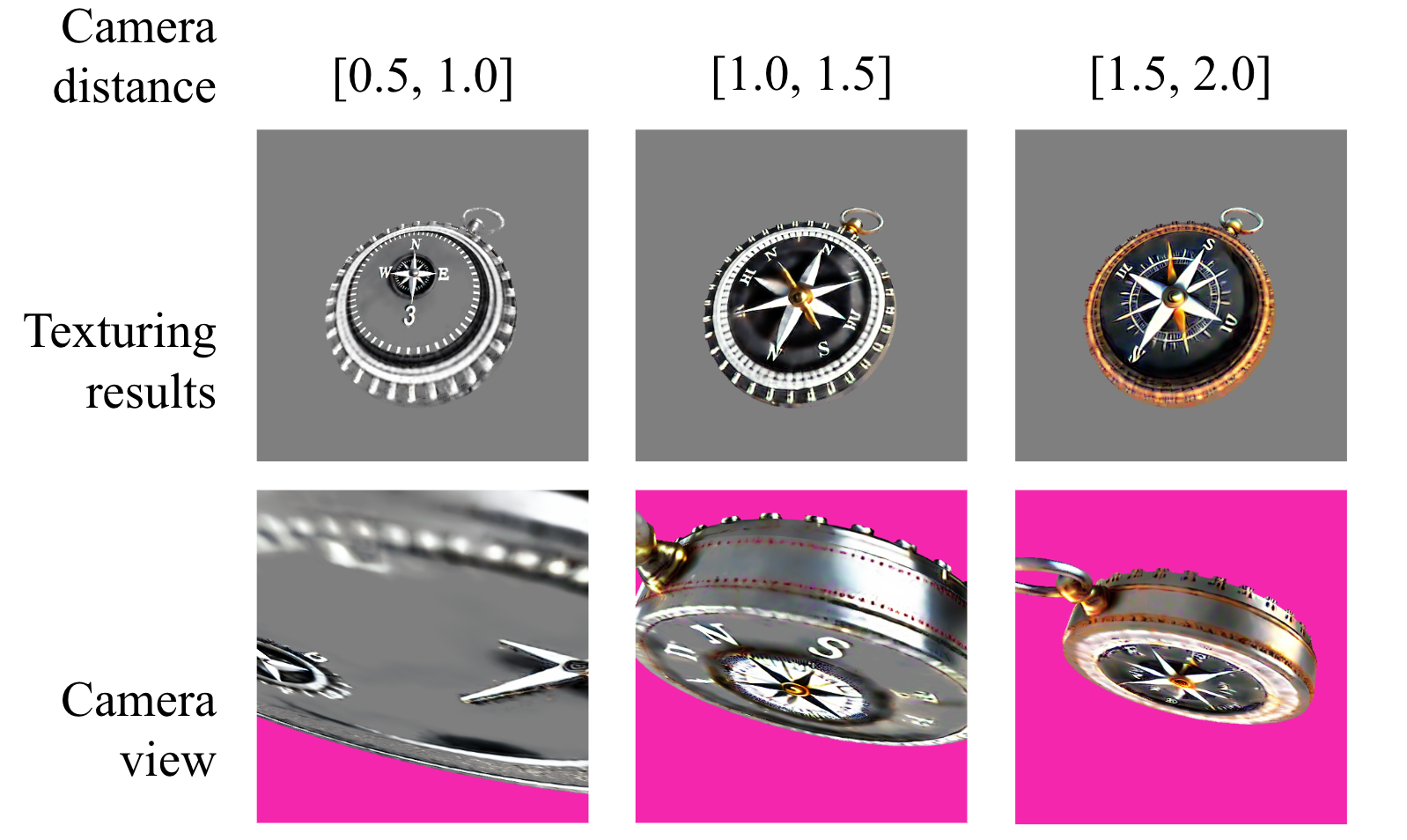}
    \caption{\textbf{Additional comparisons among camera distance values.} The one with [1.0, 1.5] has the highest quality for compass.}
    \label{fig:camera2}
\end{figure}

\begin{figure*}[h!]
    \centering
    \includegraphics[width=0.8\textwidth]{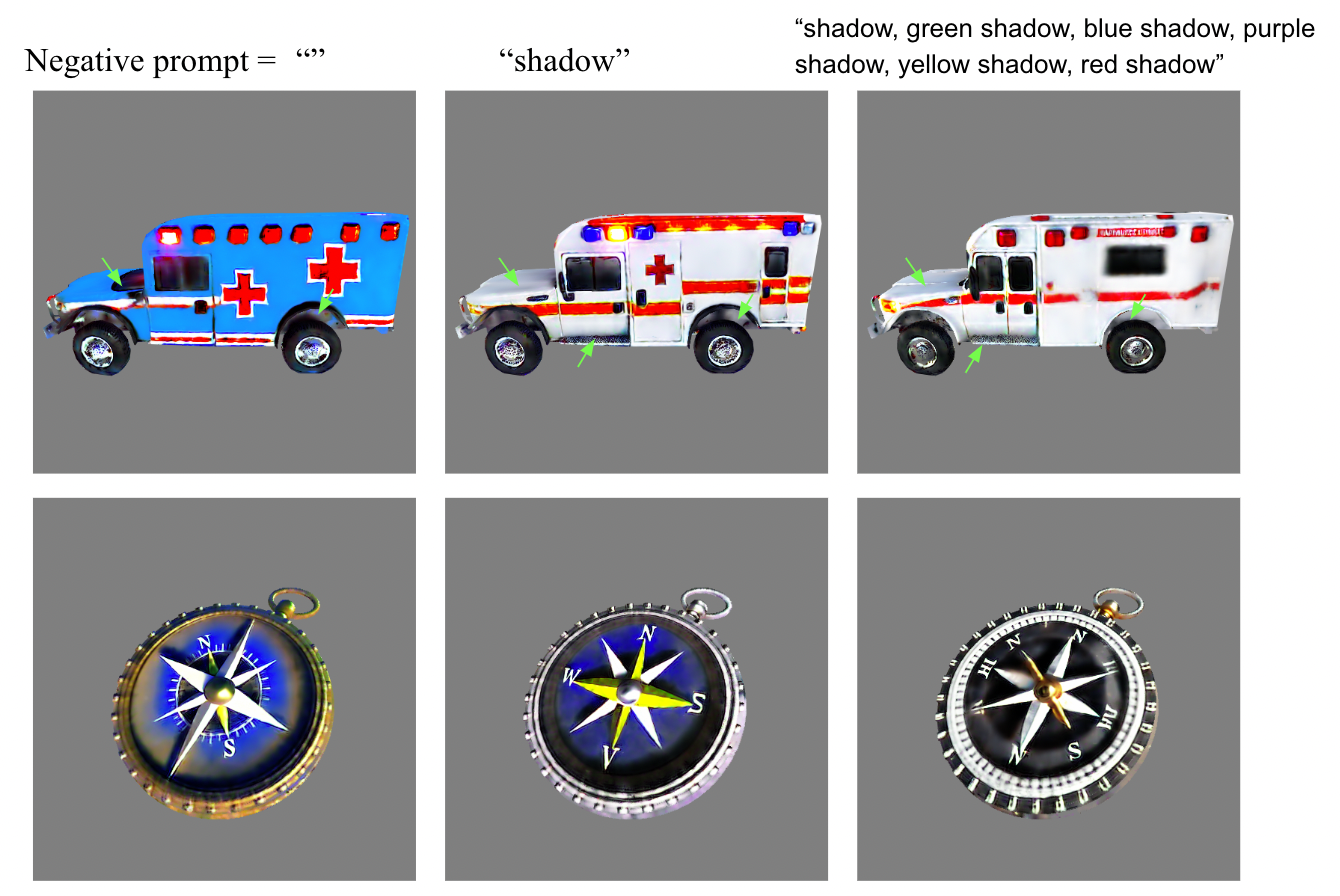}
    \caption{\textbf{Study on negative prompt.} Empty negative prompts yield weird coloring and significant baked-in shadow (see the hood). Negative prompts can be used for shadow suppression. Specifying a wide coverage removes more shadow.}
    \label{fig:neg_prompt}
\end{figure*}

\begin{figure*} [t!]
    \centering
    \includegraphics[width=0.8\textwidth]{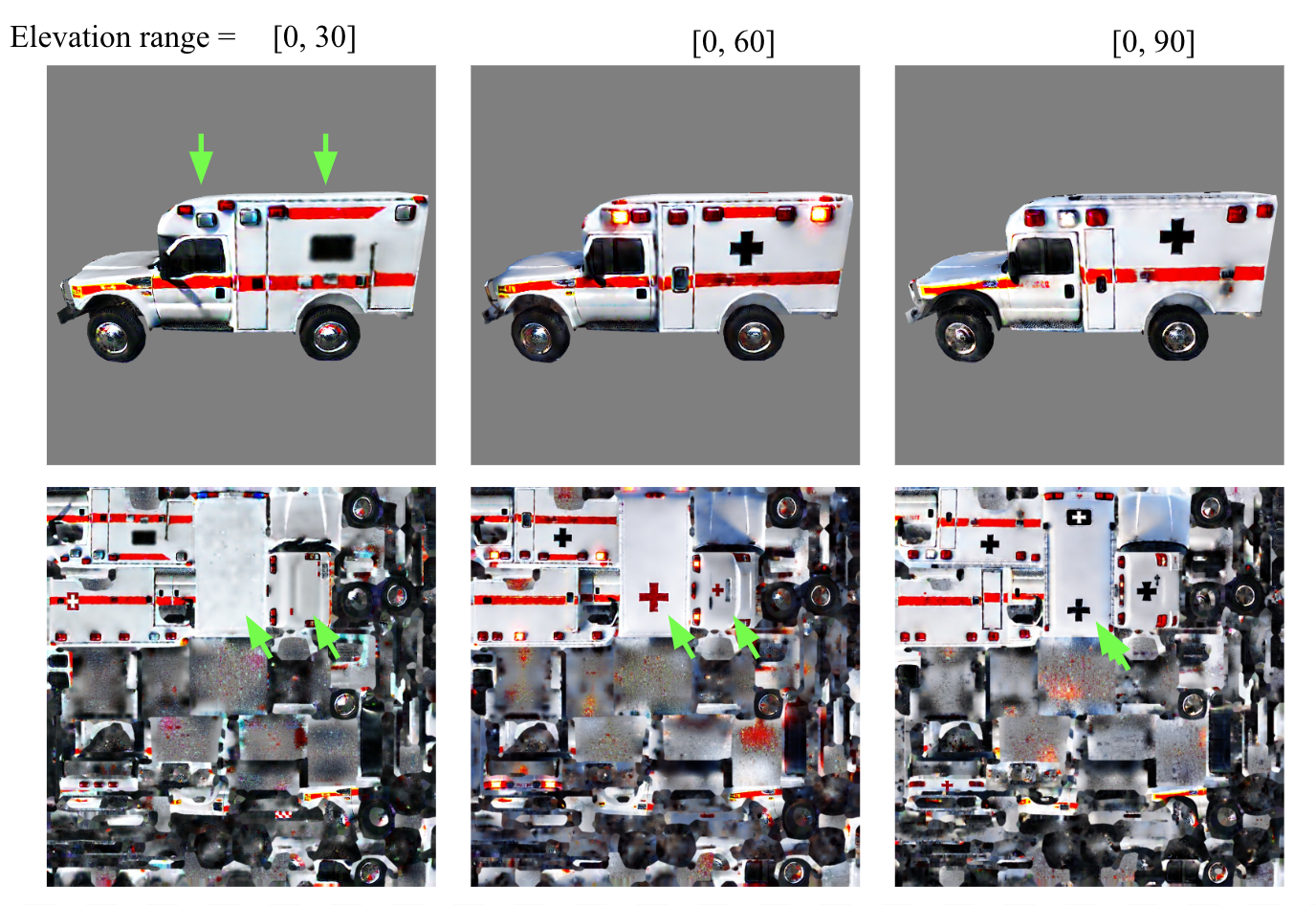}
    \caption{\textbf{Study on elevation range.} A wider camera elevation range gives better texture coverage. For example, as the elevation upper bound increases, more texture details show up on the top surfaces.
}
    \label{fig:ele_range}
\end{figure*}

\begin{figure*} [h!]
    \centering
    \includegraphics[width=0.8\textwidth]{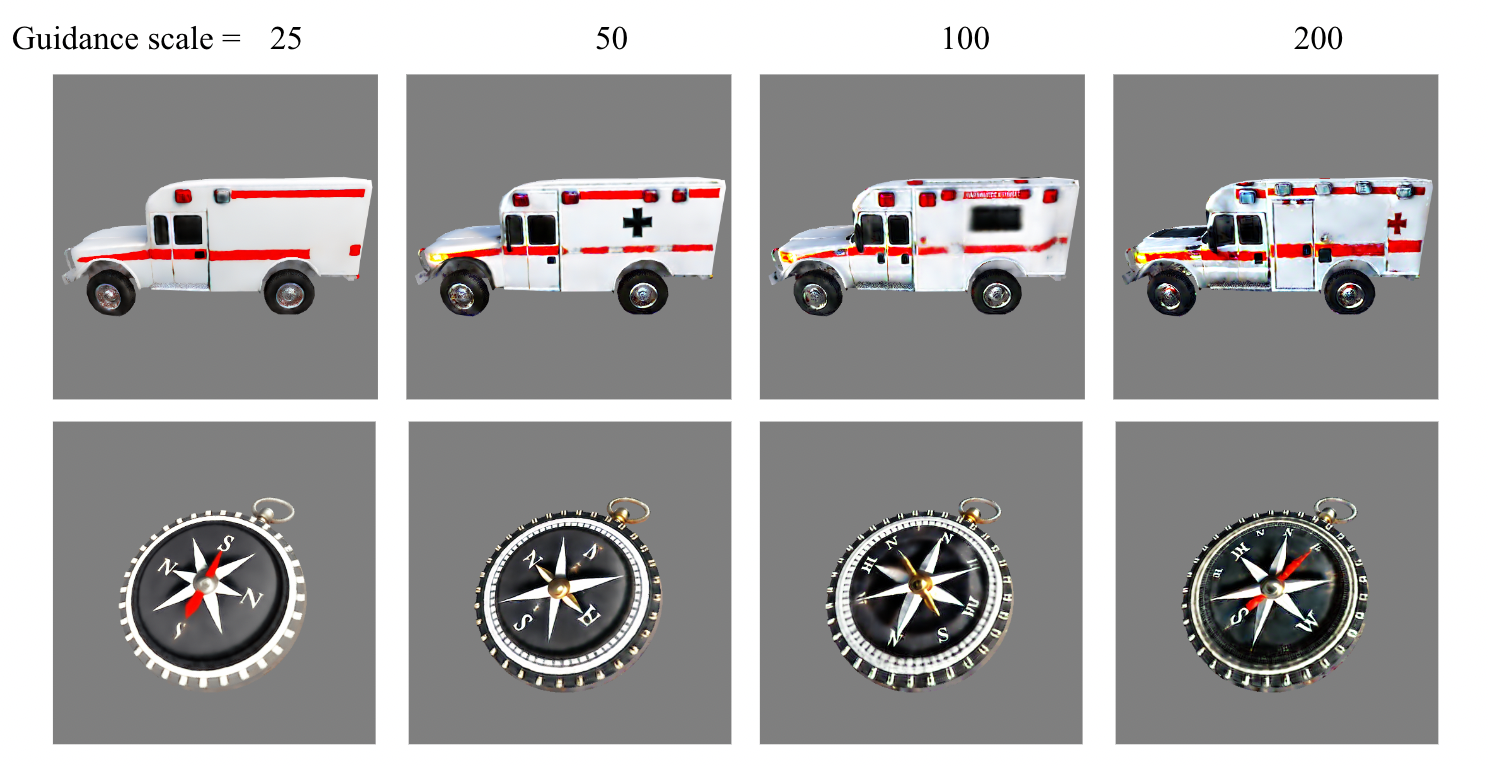}
    \caption{\textbf{Study on guidance scale.} It seems that as the guidance scale increases, more details are added, and “photorealism” seems to increase. However, it seems the noise level also increases.
}
    \label{fig:guidance}
\end{figure*}

\begin{figure*}
    \centering
    \includegraphics[width=0.8\textwidth]{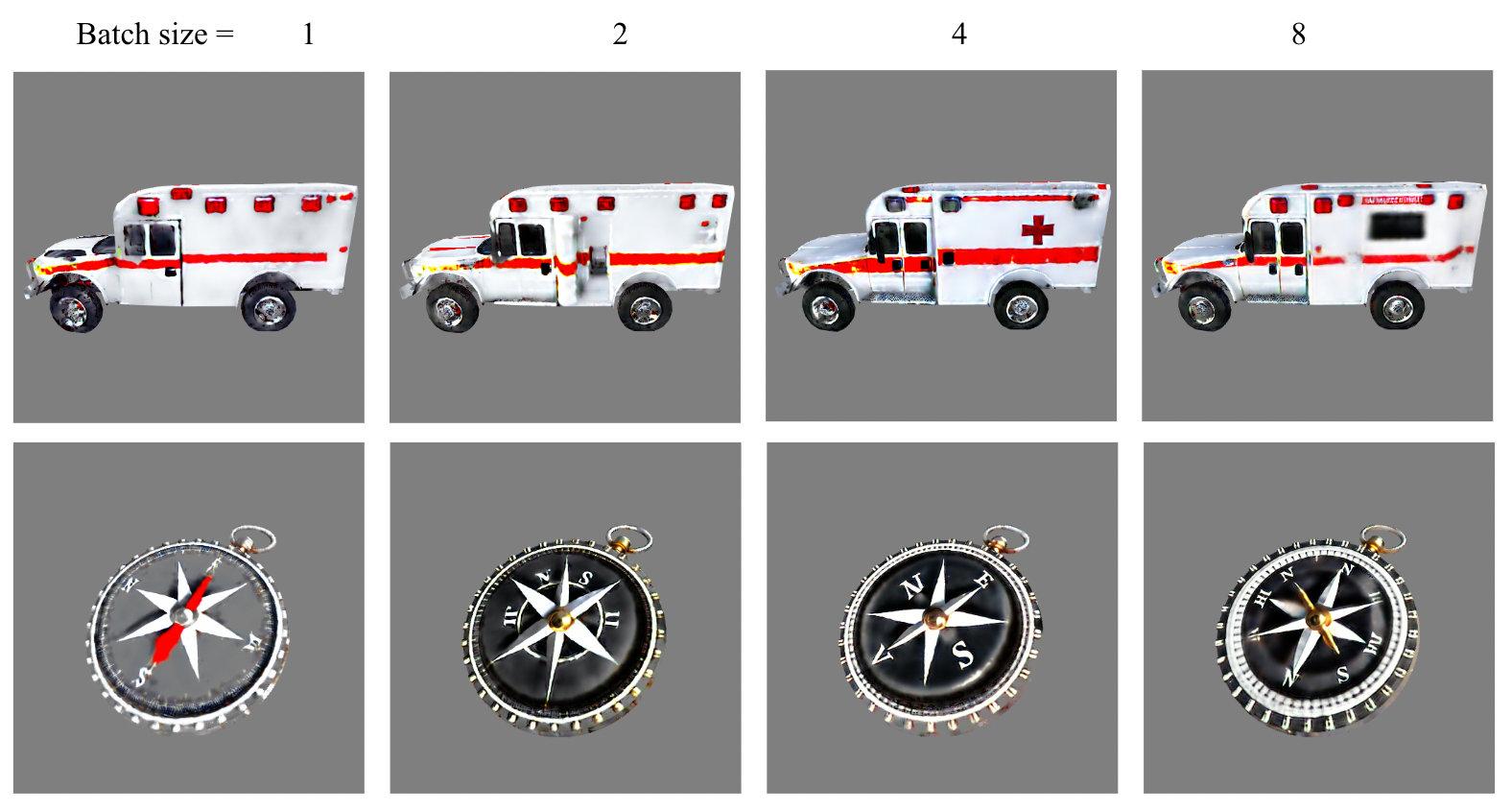}
    \caption{\textbf{Study on batch size.} Texture quality improves as batch size increases.}
    \label{fig:batch_size}
\end{figure*}

\end{document}